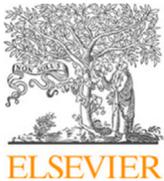
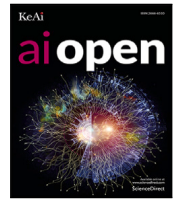

# StackVAE-G: An efficient and interpretable model for time series anomaly detection

Wenkai Li [a,d], Wenbo Hu [b,*], Ting Chen [a,*], Ning Chen [a,*], Cheng Feng [c,d]

[a] *High Performance Computing Center, Department of Computer Science and Technology, BNRist Center, Institute for AI, Tsinghua-BOSCH Joint ML Center, THBI Lab, Tsinghua University, Beijing, China*
[b] *School of Computer and Information, Hefei University of Technology, Hefei, China*
[c] *Siemens AG, Beijing, China*
[d] *THU-Siemens Joint Research Center for Industrial Intelligence and Internet of Things, Beijing, China*

A B S T R A C T

Recent studies have shown that autoencoder-based models can achieve superior performance on anomaly detection tasks due to their excellent ability to fit complex data in an unsupervised manner. In this work, we propose a novel autoencoder-based model, named StackVAE-G that can significantly bring the efficiency and interpretability to multivariate time series anomaly detection. Specifically, we utilize the similarities across the time series channels by the stacking block-wise reconstruction with a weight-sharing scheme to reduce the size of learned models and also relieve the overfitting to unknown noises in the training data. We also leverage a graph learning module to learn a sparse adjacency matrix to explicitly capture the stable interrelation structure among multiple time series channels for the interpretable pattern reconstruction of interrelated channels. Combining these two modules, we introduce the stacking block-wise VAE (variational autoencoder) with GNN (graph neural network) model for multivariate time series anomaly detection. We conduct extensive experiments on three commonly used public datasets, showing that our model achieves comparable (even better) performance with the state-of-the-art models and meanwhile requires much less computation and memory cost. Furthermore, we demonstrate that the adjacency matrix learned by our model accurately captures the interrelation among multiple channels, and can provide valuable information for failure diagnosis applications.

## 1. Introduction

Time-series anomaly detection is an important task with a wide range of applications in real-world maintenance systems (Aggarwal, 2015; Pang et al., 2020). It aims to automatically identify whether an anomaly occurs by monitoring one or multiple channels of given time-series data. Typical applications of time-series anomaly detection include intrusion detection (Portnoy et al., 2001), fraud detection (Kou et al., 2004), disease outbreak detection (Wong et al., 2003) and artificial intelligence for IT Operations (AIOps) (Lerner, 2017). For example, under the AIOps scenario, multiple sensors are deployed to monitor the operating indices of the running IT devices. These indices are supposed to be regular and predictable when devices are running normally. Time-series anomaly detection methods can be employed to learn such normal patterns and then detect the system failures automatically.

Among various solutions, autoencoder (AE) (Hinton et al., 2006) and its variants are becoming increasingly popular in recent years (Park et al., 2018; Xu et al., 2018; Su et al., 2019; Audibert et al., 2020; Xu et al., 2022) due to their strong capacity to fit complex data in an unsupervised manner. In particular, AE-based methods build a reconstruction model for the normal data and often use errors or likelihoods of the reconstructed data as the evaluation criterion for the anomaly detection task. For multivariate time series data, cross-channel similarities and interrelation structures stem from location proximity of sensors and correlations of measurements, and they widely exist in various scenarios, including moisture monitoring, health care, environmental monitoring, etc. For instance, Fig. 1 provides a typical example of real-world multivariate time series data (the raw data of the A-3 instance from the SMAP dataset (Hundman et al., 2018), which contains several telemetry indicators of the global soil moisture and freeze–thaw state). We can see the clear similarities and interrelation structures among the multiple channels. Such cross-channel structural information is valuable and can be exploited to further boost anomaly detection efficiency and interpretability of AE-based models. However, to our knowledge, existing AE-based models do not have sufficient mechanisms to capture such structural information.

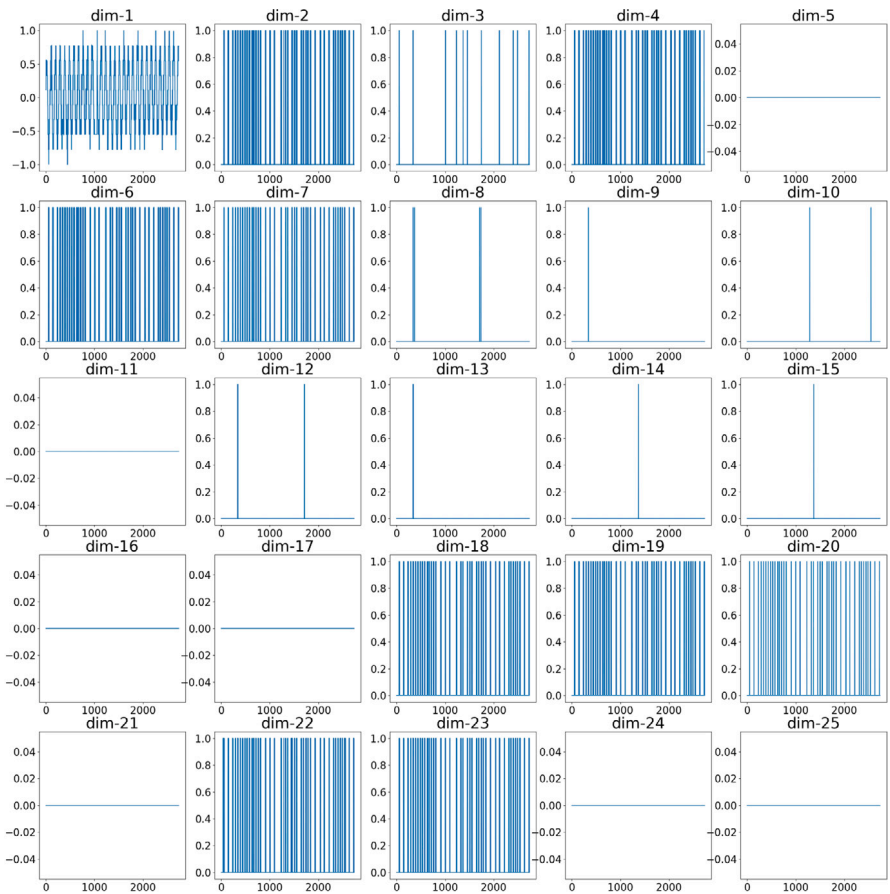

**Fig. 1.** An illustration of the 25 channels in A-3 instance of the SMAP dataset used for model training. The training set is considered to contain only normal data. From the upper left corner in the first row, these channels are indexed by channel 1, 2, 3, … The channels can be roughly clustered into several groups according to their similarity. Moreover, there are strong interrelations among channels due to the location proximity of sensors and the correlations of measurements, which will be further exploited to build our efficient and interpretable time series anomaly detection model.

For the reconstruction model, one can build a *step-wise model* with an auto-encoder (AE) (Hinton et al., 2006) or variational auto-encoder (VAE) (Kingma and Welling, 2014) unit for every time step (Park et al., 2018; Su et al., 2019; Malhotra et al., 2016) to reconstruct the series data step-by-step. A step-wise model acquires a strong expressive power from the quite large amount of model parameters, ensuring that it can fit the normal data well. However, the immoderate model capacity also makes itself easy to overfit to the noise in training data. An alternative way to reconstruct signals of multiple time steps is to compute in parallel, i.e., the *block-wise reconstruction*. By building a block-wise reconstruction model, one can efficiently obtain the multiple-time-step reconstruction through one forward pass. And the reconstruction of each step is calculated according to the whole input history, making it robust to local noise. However, the number of model parameters in a block-wise model increases dramatically when the input size grows. Thus, the block-wise model cannot handle the series patterns with long-term temporal dependencies.

To take good advantage of the merits of both *step-wise* (long-term dependency) and *block-wise* (robust reconstruction) methods, we propose the stacking block-wise framework which maintains the block-wise detection and meanwhile shares the weights among the multiple time-series channels (further illustrated in Fig. 2). One merit of such framework is that it incorporates the cross-channel similarities to build a light, efficient model using a channel-wise weight-sharing scheme, which relieves overfitting while enables block-wise reconstruction for long sliding windows. Moreover, this framework enables AE-based models to incorporate structural information for multivariate time series anomaly detection. We further leverage a graph learning module under such framework to learn the interrelation structure information via a sparse adjacency matrix for the time series channels. The sparse adjacency matrix is learned via a self-supervised graph learning loss and a top-k neighbourhood setting. It is designed to deliver interpretable information for series reconstruction and the fault diagnosis application.

We conduct extensive experiments on three commonly used multivariate time series anomaly detection datasets in the recent literature. Experimental results show that our method not only achieves the superior time-series anomaly detection performance with efficiency compared with several start-of-the-art baseline methods, but also learns the interpretable structural information via the graph learning module. Moreover, we also show that our model requires much less memory and computation cost compared with other start-of-the-art AE-based time series anomaly detection models, making our model favourable for deployment in resource-limited environments like edge devices.

In summary, our contributions are as follows:

- We propose a stacking block-wise reconstruction framework with weight-sharing for auto-encoder-based anomaly detection models which is significantly efficient compared with the previous step-wise and block-wise models.
- We incorporate a graph learning module to learn the interrelation structure in the stacking block-wise model, which brings interpretability for the time series anomaly detection task.
- We conduct quantitative and qualitative experiments to show that our method has competitive detection results with high efficiency and interpretability.

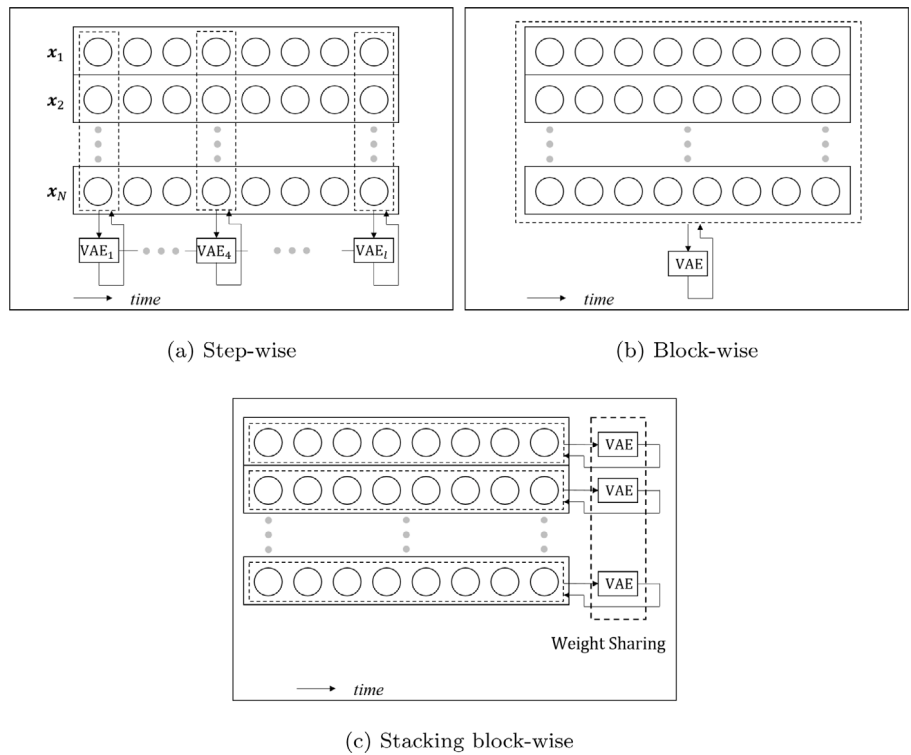

**Fig. 2.** The architecture of AE-based models for times-series anomaly detection. Figs. 2(a) and 2(b) are for previous works. Fig. 2(c) is proposed in this paper. Here we use VAE as a representative to show its possible usage when reconstructing one slice $X \in \mathbb{R}^{N \times l}$.

## 2. Background

In this section, we provide the background of time-series anomaly detection and AE-based models. We also introduce the related methods of graph learning and structure learning methods for time-series data.

### 2.1. Time-series anomaly detection

Given a multivariate time-series dataset $S \in \mathbb{R}^{N \times T}$, we denote the number of the channels as $N$ and the length of the time-series as $T$. The goal of time-series anomaly detection is to find the anomalous points and segments among these $T$ time steps. In practice, we use a sliding window to split the original multivariate time-series $S \in \mathbb{R}^{N \times T}$ into multiple slices with a pre-defined window size $l$: $X_l, X_{l+1}, \ldots, X_t, \ldots, X_T$. The slice $X_t$ contains $l$ multivariate observations ranging from the time step $t - l + 1$ to $t$. Although time-series data may have various types of anomalies, the anomalies can be detected by exploiting the normal patterns such as trend and seasonality (Aggarwal, 2015).

If there is only one dimension for the time-series data $S$, i.e., $N = 1$, we call it univariate (or single-channel) time-series anomaly detection, and the anomaly detection models mainly explore the temporal dependency to identify abnormal patterns. When it comes to the multivariate data, i.e., $N > 1$, besides the temporal dependency of the time-series, we should also explore the channel-level interrelations, which usually remain stable when the monitored system is running normally, as the additional structure information for solving the anomaly detection task.

Supervised methods, such as deep neural networks and support vector machines, can be directly migrated to time-series anomaly detection by considering it as a classification task (Aggarwal, 2015). However, in practice the label scarcity and diversity of time-series anomalies prevent further applications of such supervised methods. Unsupervised anomaly detection methods are more commonly used than supervised ones. Traditional unsupervised anomaly detection methods, such as one-class SVM (Manevitz and Yousef, 2001) and isolation forest (Liu et al., 2008), can be applied to the time-series data with temporal information ignored or implicitly captured. Originated from these traditional models, several deep models based on clustering (Ruff et al., 2018; Shin et al., 2020; Shen et al., 2020) are proposed. In particular, DeepSVDD (Ruff et al., 2018) squeezes the deep latent feature of normal samples into one small hypersphere; ITAD (Shin et al., 2020) decomposes the constructed tensors and performs clustering on one component; and THOC (Shen et al., 2020) proposes a temporal hierarchical clustering mechanism under the one-class classification setting. Apart from the aforementioned traditional and clustering-based methods, generally, the existing unsupervised approaches based on neural networks can be classified into prediction-based methods and reconstruction-based methods. As an example of prediction-based methods, LSTM-NDT (Hundman et al., 2018) firstly uses LSTM to perform time series forecasting and on this basis the nonparametric dynamic thresholding method is used on prediction errors to detect anomalies. For reconstruction-based methods, AE-based models have been widely adopted with fruitful progress, which is detailed in the following part.

### 2.2. Autoencode-based models for time-series anomaly detection

AE-based models are a class of the most popular methods for the time series anomaly detection task. Generally, such models reconstruct the normal patterns of the raw input data and detect anomalies in an unsupervised fashion by using errors or reconstruction likelihoods as anomaly scores (Aggarwal, 2015).

As stated in Section 1, there exists two general categories of AE-based methods for time series anomaly detection — step-wise and block-wise methods, as illustrated in Fig. 2(a) and Fig. 2(b) respectively. In detail, most *step-wise approaches* basically use a recurrent neural network (RNN) cell to build the reconstruction model at each time step. For instance, LSTM-Encoder-Decoder (Malhotra et al., 2016) uses an LSTM cell to encode the multivariate time-series for each time step and then reconstructs the input series reversely in the autoencoder framework. LSTM-VAE (Park et al., 2018) equips the LSTM cell in the VAE framework, showing VAE's superiority of the probabilistic reconstruction in the continuous latent space. Then GGM-VAE (Guo et al., 2018) assumes a Gaussian mixture prior in the latent space to replace the uni-modal Gaussian assumption of VAE. OmniAnomaly (Su et al., 2019) employs a stochastic recurrent neural network with stochastic variable connection and planar Normalizing Flows (Rezende and Mohamed, 2015) to further describe non-Gaussian distributions of latent space. Anomaly Transformer (Xu et al., 2022) renovates the backbone of the transformer (Vaswani et al., 2017) with an Anomaly-Attention mechanism to capture the association discrepancy for better distinguishability on abnormal samples. However, one common weakness of the step-wise methods is that they are prone to overfit the noise widely existing in time series data due to the strategy of building a reconstruction model on every time step.

*Block-wise approaches* choose to reconstruct the time-series data of sliding windows block by block. For instance, Donut (Xu et al., 2018) is a single-channel anomaly detection model based on VAE with a modified ELBO (M-ELBO), missing data injection, and MCMC imputation techniques. Audibert et al. (2020) proposed the UnSupervised Anomaly Detection (USAD) for multivariate time-series, which introduces adversarial training into the training procedure of a block-wise reconstruction model. The adversarial training amplifies the reconstruction error of inputs so that the USAD model can detect the anomalies with small deviates. As the amount of model parameters in block-wise models increases dramatically when the input size grows (will be shown in experiments), for time series with multiple channels, they could only choose a small window size to avoid the huge amount of model parameters. As a result, they are unable to capture long-term temporal dependencies.

*2.3. Graph learning and structure learning for time-series*

Regarding modelling multivariate time-series data, one can use auto-regression methods with other endogenous variables and build up a vector auto-regression (VAR) model (Hamilton, 2020). Recurrent neural networks, such as LSTM (Hochreiter and Schmidhuber, 1997), can be used to build deep learning models for multivariate time-series data. But in these recurrent deep models (also including the step-wise approaches (Malhotra et al., 2016; Park et al., 2018; Su et al., 2019)), the relationships among the channels are implicitly represented in the deep neural networks, resulting in a lack of interpretability. GGM-VAE (Guo et al., 2018) improves the model interpretability by assuming a Gaussian mixture prior in the latent space, but the structural information in the latent mixture space cannot correspond to the channel dimension.

After evaluating the existing time-series anomaly detection models from the structure learning perspective, we briefly introduce the specialized structure learning algorithms on multivariate time-series data. The prior knowledge for the structural information of multivariate time-series is often unknown, so it is a challenge to model the interrelation among the multiple channels. This interrelation structural information can be derived from the feature sparsity and represented as graphs (Hui, 2006; Bach and Jordan, 2004). Graphical lasso (Friedman et al., 2007) transferred the structure learning problem to inverse covariance matrix estimation. More recently, Wu et al. (2020) proposed a graph neural network module for learning the directed spatial dependency for the multivariate time-series forecasting tasks. Other types of time series interpretability includes the saliency map for forecasting (Pan et al., 2021) and the Granger causality among the multiple time series (Zhang et al., 2020).

## 3. Methodology

We now present our proposed Stacking-VAE with Graph neural network (StackVAE-G) model, which consists of two modules — a stacking block-wise VAE (StackVAE) with a weight-sharing scheme and a graph neural network (GNN) module.

Fig. 3 shows the overall architecture of our model. In order to be effective on anomaly detection, StackVAE builds the reconstruction model by stacking single-channel reconstruction procedures with shared weights to exploit the channel-level similarities within multivariate time series data.

*3.1. Variational auto encoder*

We briefly introduce the variational auto-encoder (VAE) (Kingma and Welling, 2014), which is the base model of StackVAE. VAE is a deep generative model that characterizes the relationship between the input data sample $x$ and its corresponding latent code $z$. Usually we assign a standard Gaussian distribution $p(z) = \mathcal{N}(\mathbf{0}, \mathbf{I})$ to the latent code $z$ as its prior distribution. A VAE model is composed of two parts — a decoder network $p(x|z)$ for generating observations by conditioning on the latent code and an encoder network $q(z|x)$ for inferring the distribution of latent code for each individual input. In VAE, the encoder network is needed as an approximation to the intractable posterior distribution $p(z|x)$ under the framework of variation inference, and it is usually assumed to be multivariate Gaussian distribution $q(z|x) = \mathcal{N}(\mu_z, \sigma_z^2 \mathbf{I})$. Then, the parameters of the encoder and decoder networks are jointly optimized by maximizing the evidence lower bound (ELBO):

$$J = \mathbb{E}_{q(z|x)} \left[ \log p(x|z) \right] - \mathrm{KL}(q(z|x) \parallel p(z)). \tag{1}$$

The common strategy of stochastic gradient descent (SGD) has been extended to solve this problem with promising performance.

*3.2. Stacking block-wise VAE model with a weight-sharing scheme*

The stacking block-wise VAE reconstruction model builds a single-channel block-wise reconstruction and stacks it multiple times with weight-sharing as illustrated in Fig. 2(c). Specifically, we define the input slice $X = \left[ x_1, \ldots, x_N \right]^\top$ and the reconstruction output $\hat{X} = \left[ \hat{x}_1, \ldots, \hat{x}_N \right]^\top$, where $x_n$ and $\hat{x}_n$ correspond to the series data of channel $n$. The latent code of $x_n$ is defined as $z_n$. By stacking $z_n$ of all channels using the same model, we get the latent code $Z = \left[ z_1, \ldots, z_N \right]^\top$ of the slice $X$.

The encoding process of our model maps the slice $X$ to its posterior latent distribution $q(Z|X)$. For each channel in a slice, $x_n \in \mathbb{R}^l$ (one row of $X \in \mathbb{R}^{N \times l}$), the encoder of the standard VAE module maps $x_n$ into the corresponding posterior Gaussian distribution:

$$q(z_n | x_n) = \mathcal{N}(\mu_{z_n}, \sigma_{z_n}^2 \mathbf{I}), \quad n = 1, \ldots, N, \tag{2}$$

where $z_n$ has $m$ dimensions, $\mu_{z_n}$ and $\sigma_{z_n}$ represent the means and standard deviations of $m$ independent Gaussian components in $z_n$ and $\mathbf{I}$ is an identity matrix. Thus the approximated posterior latent distribution for the slice $X$, i.e., $q(Z|X)$, is obtained with each component $q(z_n|x_n)$ identified. Then we randomly sample from $q(z_n|x_n)$ to get the latent code $z_n$ for each channel. The complete encoding procedure is formulated as follows:

$$\text{for } n = 1, \ldots, N : \quad h_{n,1} = ReLU \left[ f(x_n) \right],$$
$$H_1 = \left[ h_{1,1}, \ldots, h_{N,1} \right]^\top,$$
$$H_2 = (1-\gamma) H_1 + \gamma \widetilde{A} \times H_1,$$
$$\text{for } n = 1, \ldots, N : \quad \begin{cases} \mu_{z_n} = f(h_{n,2}), \\ \sigma_{z_n} = \text{SoftPlus} \left[ f(h_{n,2}) \right], \end{cases} \tag{3}$$

where $f$ is a neural linear layer shared among different channels, $h_{n,2}^\top$ is a row of $H_2$ (i.e., $H_2 = \left[ h_{1,2}, \ldots, h_{N,2} \right]^\top$), and $\widetilde{A}$ is a sparse channel-interrelation square matrix obtained by the graph learning module, which will be introduced in the next subsection.

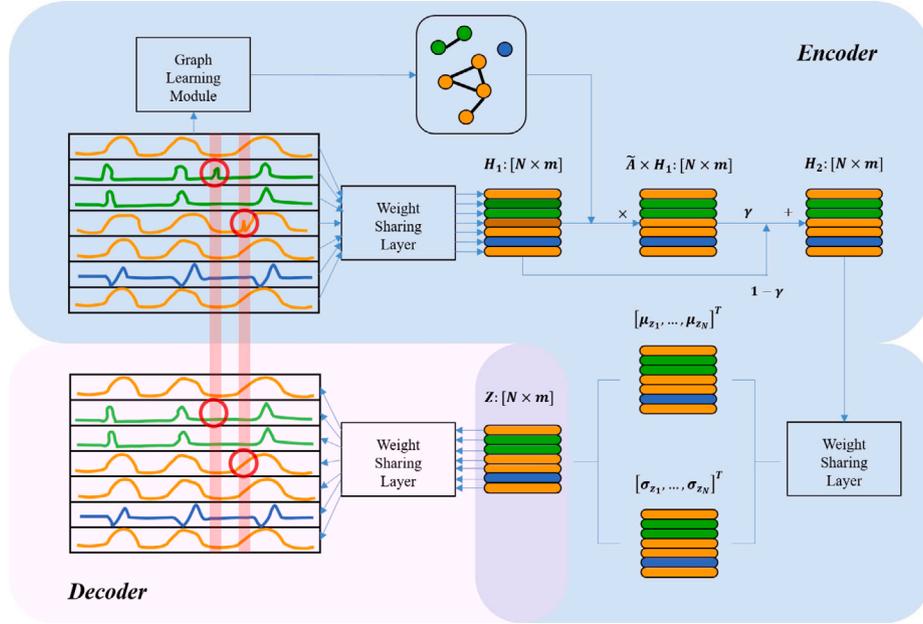

**Fig. 3.** The architecture of StackVAE-G. Under the stacking block-wise reconstruction framework, the multivariate time-series data is encoded into the mid-latent feature $H_1$. Then the single-channel VAE model is stacked with a weight-sharing Linear layer for multivariate time series data with channel-level similarities. Moreover, the channels are regarded as graph nodes and a graph neural network module is used to learn the interrelation among the channels.

The encoding procedure aims to recognize the normal patterns in $X$, thus to infer the corresponding normal latent code $Z$. To get rid of the inference bias introduced by anomalies in $X$, we combine $\widetilde{A}$ with $H_1$ instead. With the neighbours and corresponding interrelation coefficients provided in $\widetilde{A}$, the intermediate latent feature of channel $n$: $h_{n,2}$ retains its original state $h_{n,1}$ with the ratio $1 - \gamma$, while shares the information propagated from its neighbours with the ratio $\gamma$. If $\gamma$ is set as 0, StackVAE-G will degenerate to the model without the graph learning module, StackVAE. In such an architecture design, StackVAE adopts a more robust block-wise model and reduces the model size via the weight sharing technique. As we shall see in experiments, Stack-VAE derives an effective reconstruction for the time series anomaly detection problem.

The decoder works as the inverse procedure of the encoder, that is, for $n = 1, \ldots, N$:

$$\begin{aligned}
\hat{\boldsymbol{h}}_n &= ReLU\left[f(\boldsymbol{z}_n)\right], \\
\boldsymbol{\mu}_{\hat{\boldsymbol{x}}_n} &= f(\hat{\boldsymbol{h}}_n), \\
\boldsymbol{\sigma}_{\hat{\boldsymbol{x}}_n} &= \text{SoftPlus}\left[f(\hat{\boldsymbol{h}}_n)\right]
\end{aligned} \quad (4)$$

where $\hat{\boldsymbol{h}}_n$ denotes the intermediate latent feature, and $\boldsymbol{\mu}_{\hat{\boldsymbol{x}}_n}$ and $\boldsymbol{\sigma}_{\hat{\boldsymbol{x}}_n}$ represent the mean and variance of the normal distribution for generating $\boldsymbol{x}_n$, respectively:

$$p(\boldsymbol{x}_n|\boldsymbol{z}_n) = \mathcal{N}(\boldsymbol{\mu}_{\hat{\boldsymbol{x}}_n}, \boldsymbol{\sigma}_{\hat{\boldsymbol{x}}_n}^2 \mathbf{I}). \quad (5)$$

With the individual components $p(\boldsymbol{x}_n|\boldsymbol{z}_n)$ for channel $n$, we can get the overall generating distribution $p(X|Z)$ by multiplying all components together.

In practice, $\boldsymbol{\mu}_{\hat{\boldsymbol{x}}_n}$ can be adopted as the reconstruction output $\hat{\boldsymbol{x}}_n$. By stacking $\hat{\boldsymbol{x}}_n$, we get the reconstruction output $\hat{X} = \left[\hat{\boldsymbol{x}}_1, \ldots, \hat{\boldsymbol{x}}_N\right]^\top$.

### 3.3. Graph learning module

The graph learning module is designed to characterize the interrelation structure among multiple series channels via the adjacency matrix $\widetilde{A}$ in Eq. (3). Specifically, we regard the channels as the nodes and learn an undirected graph from the representation of each channel data via the following graph neural networks:

$$M = \tanh(\alpha f(E)),$$

$$A = ReLU(\alpha \tanh(MM^\top)), \quad (6)$$

where $E$ denotes the randomly initialized node embedding for channels and $\alpha$ is an amplifier parameter that controls the saturation rate of the activation function ($\tanh$ and $ReLU$).

To ensure the sparsity of the adjacency matrix $A$, we retain the top-k values per row and set the others to be zeros. The tailored $\widetilde{A}$ is formulated as follows: for $n = 1, 2, \ldots, N$:

$$\begin{aligned}
idx &= \text{argtopk}(A[n, :]) \\
A[n, -idx] &= 0 \\
D &= \text{diag}\left[1 + \sum_{j=1}^{N} A_{1j}, \ldots, 1 + \sum_{j=1}^{N} A_{Nj}\right] \\
\widetilde{A} &= D^{-1}(I + A)
\end{aligned} \quad (7)$$

where the function argtopk returns the indexes of the top-k values in a 1-D tensor, $-idx$ represents the complement set of $idx$, containing the other $N - k$ values and diag converts a $1 \times N$ tensor into an $N \times N$ diagonal matrix. Importantly, we propose a self-supervised regression-based graph learning loss for guiding the graph structure learning, which aims to discover the *time-invariant* interrelation structure of series channels:

$$\begin{aligned}
\mathcal{L}_{\text{Graph}} &= \|X - \widetilde{A} \times X\|_2^2 \\
&= \sum_{n=1}^{N} \|\boldsymbol{x}_n - \sum_{j=1}^{N} \widetilde{A}_{nj} \boldsymbol{x}_j\|_2^2
\end{aligned} \quad (8)$$

where $X$ represents the input slice and $\widetilde{A}$ is the tailored adjacency matrix. Each row in $\widetilde{A}$ contains time-invariant weights for neighbours, according to which each channel can reconstruct itself by inspecting its neighbours through linear regression (Eq. (8)). The adjacency matrix $\widetilde{A}$ is independent of time under the guidance of $\mathcal{L}_{\text{Graph}}$, which describes the stable linear dependencies along the channel dimension.

In Eq. (3), the mid-latent feature $H_1$ extracts the temporal information individually for each series channel. No channel-level interaction is involved in this stage. With the learned $\widetilde{A}$, it is natural and convenient to fuse the channel interrelation into the latent code $H_2$ through simple matrix multiply. $H_2$ combines mid-latent signals in $H_1$ using weights provided by $\widetilde{A}$. Thus the latent code $\boldsymbol{z}_n$ is no more determined by $\boldsymbol{x}_n$ only. Signals from its highly-related neighbour channels are also taken into consideration to infer a reliable latent code $z_n$.

The core innovation of the graph learning module lies in the self-supervised loss design. Under the regression-based graph loss $\mathcal{L}_{\text{Graph}}$, each item in $\widetilde{A}$ explicitly represents a regression coefficient. Thus $\widetilde{A}$ becomes an interpretable weight matrix and is easily adopted in downstream tasks (aforementioned latent code inference and anomaly diagnosis). We also conduct the ablation study on using the graph learning loss or not (Wu et al., 2020) in Section 4.6, which empirically shows its impact to learn meaningful interrelation structure among channels for time series anomaly detection tasks.

### 3.4. Loss function

The learning objective of StackVAE is the same as that of VAE, that is to maximize the evidence lower bound (ELBO), which is equivalent to minimizing the following loss function:

$$\mathcal{L}_{\text{VAE}} = -\mathbb{E}_{q(Z|X)}[\log p(X|Z)] + \text{KL}(q(Z|X) \parallel p(Z)). \quad (9)$$

Combining with the graph learning loss introduced in Section 3.3, the overall loss function of StackVAE-G is given as follows:

$$\begin{aligned}\mathcal{L}_{\text{total}} &= \mathcal{L}_{\text{VAE}} + \lambda \mathcal{L}_{\text{Graph}} \\ &= -\mathbb{E}_{q(Z|X)}[\log p(X|Z)] + \text{KL}(q(Z|X) \parallel p(Z)) \\ &\quad + \lambda \|X - \widetilde{A} \times X\|_2^2 \end{aligned} \quad (10)$$

where $\lambda$ is a hyper-parameter to balance the two parts of the loss function.

### 3.5. Training & detecting procedures

For training, the two modules of StackVAE-G, namely the StackVAE model and the graph learning module, are jointly trained by optimizing $\mathcal{L}_{\text{total}}$ in Eq. (10) using Adam (Kingma and Ba, 2015). The encoder of StackVAE together with the graph learning module are trained to infer the true posterior latent distribution $p(X|Z)$ with interpretable interrelation among channels. After random sampling from the approximated posterior $q(Z|X)$, the decoder is trained to accurately generate the reconstruction outputs $\hat{X}$. After training for several epochs, the random sampling process guarantees the continuous space formed by these $Z$ samples is linked to the normal pattern of input $X$.

For detection, with the reconstructed normal pattern $\hat{X} = [\hat{x}_1, \ldots, \hat{x}_N]^\top$ for input $X$, the anomaly score of each time step $t$ in $X$ is defined as the sum of the squared error between original inputs and reconstruction outputs:

$$s_t = \sum_{n=1}^{N} (\hat{x}_{n,t} - x_{n,t})^2 \quad (11)$$

where $x_{n,t}$ represents the value of $\hat{x}_n$ at time step $t$ and $\hat{x}_{n,t}$ is the reconstruction value of $x_{n,t}$. After getting the anomaly scores, we select a constant threshold $c$. Then the time step $t$ with $s_t > c$ will be considered as an anomalous point. Though we may calculate the score for each time step in one slice $X$, when performing online detection we only use the score of the last point for a quick real-time response. For the detection outputs, anomalies often occur in the form of continuous anomalous segments. And once an alert is triggered within the lasting period of an anomaly, this anomalous segment is considered to be detected. We adopt the point-adjust evaluation approach (Xu et al., 2018) which recognize a correct detection if any point in this segment is detected.

## 4. Experiments

We now present the experimental results of StackVAE-G on various datasets by comparing with state-of-the-art baselines. We also conduct insightful qualitative results to show the efficiency and interpretability of our method.

### 4.1. Data and evaluation metrics

We conduct our experiments on three frequently used public datasets, which is consistent with the experiment settings of the recent state-of-the-art methods (Su et al., 2019; Audibert et al., 2020; Xu et al., 2022). The details are as follows:

- Server Machine Dataset (SMD): SMD is published by Su et al. (2019), which lasts for 5 weeks long, monitors 28 server machines for a large Internet company with 33 sensors. SMD is one of the largest public datasets currently available for evaluating multivariate time-series anomaly detection. It contains metrics like CPU load, network usage, memory usage, etc.
- Soil Moisture Active Passive (SMAP) satellite and Mars Science Laboratory (MSL) rover Datasets: SMAP and MSL are two datasets published by NASA (Hundman et al., 2018). SMAP contains 55 entities and each entity is monitored by 25 sensors. MSL contains 27 entities and 66 sensors for each entity. The metrics in the two datasets include telemetry data, radiation, temperature, power, computational activities, etc (Su et al., 2019).

We use *precision* (P), *recall* (R), and *F1 score* (F1) as the evaluation metrics.

### 4.2. Baseline methods

StackVAE-G is the proposed method to be evaluated and we also include the following variations into comparisons:

- *StackVAE*: the StackVAE-G model with $\gamma = 0$ (i.e., omitting the graph learning module);
- *StackVAE+Fixed Graph*: a two-stage model which firstly learns the graph structure with the lasso regression and then uses the fixed structure in the stacking block-wise reconstruction framework for detection.

We select a wide range of 11 representative unsupervised models for multivariate time-series anomaly detection as our baselines, including OC-SVM (Manevitz and Yousef, 2001), Isolation Forest (Liu et al., 2008) (IF), DeepSVDD (Ruff et al., 2018), LSTM-VAE (Park et al., 2018), DAGMM (Zong et al., 2018), LSTM-NDT (Hundman et al., 2018), OmniAnomaly (Su et al., 2019), USAD (Audibert et al., 2020), ITAD (Shin et al., 2020), THOC (Shen et al., 2020) and Anomaly Transformer (Xu et al., 2022). For each model, we test its possible anomaly thresholds and report its best performance according to the metric of F1-score, which is consistent with (Audibert et al., 2020; Xu et al., 2022).

We use the lasso regression for each channel as a qualitative baseline to compare the learned interrelation graph structure. Let the data of channel $n$: $x_n$ be the target and the other channels $x_{-n} = [x_1, x_2, \ldots, x_{n-1}, x_{n+1}, \ldots, x_N]^\top$ be the predictors. The lasso regression aims to find the regression coefficients $\beta_n \in \mathbb{R}^{N-1}$:

$$\beta_n = \arg\min_{\beta} \left\{ \mathbb{E}_t \left[ \frac{1}{2} |\beta_n^\top x_{-n,t} - x_{n,t}|^2 \right] + \lambda \|\beta_n\|_1 \right\}. \quad (12)$$

We solve Eq. (12) for each channel from $x_1$ to $x_N$. Note that $\beta_n$ contains $N-1$ coefficients for $x_{-n}$ excluding $x_n$ itself, it is equivalent to add $x_n$ as a predictor with the regression coefficient 0. Thus we obtain $\hat{\beta}_n \in \mathbb{R}^N$ with $\hat{\beta}_{n,n} = 0$. Finally, we learn the regression coefficient matrix $A = [\hat{\beta}_1, \hat{\beta}_2, \ldots, \hat{\beta}_N]^\top$. Following the process in Eq. (7), we calculate $D$ and then get the final output $\widetilde{A} = D^{-1}(I + A)$.

### 4.3. Hyper-parameters for stackvae-g

There are four hyper-parameters for the graph learning module, which are chosen via a grid search:

**Table 1**
Performance comparison.

| Methods | SMD | | | SMAP | | | MSL | | |
|---|---|---|---|---|---|---|---|---|---|
| | P | R | F1 | P | R | F1 | P | R | F1 |
| OC-SVM | 0.4434 | 0.7672 | 0.5619 | 0.5385 | 0.5907 | 0.5634 | 0.5978 | 0.8687 | 0.7082 |
| IF | 0.5938 | 0.8532 | 0.7003 | 0.4423 | 0.5105 | 0.4739 | 0.5681 | 0.6740 | 0.6166 |
| DeepSVDD | 0.7854 | 0.7967 | 0.7910 | 0.8993 | 0.5602 | 0.6904 | 0.9192 | 0.7663 | 0.8358 |
| LSTM-VAE | 0.8698 | 0.7879 | 0.8268 | 0.7164 | 0.9875 | 0.8304 | 0.8599 | 0.9756 | 0.9141 |
| DAGMM | 0.6730 | 0.8450 | 0.7493 | 0.6334 | 0.9984 | 0.7751 | 0.7562 | 0.9803 | 0.8537 |
| LSTM-NDT | 0.5684 | 0.6438 | 0.6037 | 0.8965 | 0.8846 | 0.8905 | 0.5934 | 0.5374 | 0.5640 |
| OmniAnomaly | 0.9809 | 0.9438 | **0.9620** | 0.7585 | 0.9756 | 0.8535 | 0.9140 | 0.8891 | 0.9014 |
| USAD | 0.9314 | 0.9617 | 0.9463 | 0.7697 | 0.9831 | 0.8634 | 0.8810 | 0.9786 | 0.9272 |
| ITAD | 0.8622 | 0.7371 | 0.7948 | 0.8242 | 0.6689 | 0.7385 | 0.6944 | 0.8409 | 0.7607 |
| THOC | 0.7976 | 0.9095 | 0.8499 | 0.9206 | 0.8934 | 0.9068 | 0.8845 | 0.9097 | 0.8969 |
| Anomaly-Transformer | 0.8940 | 0.9545 | 0.9233 | 0.9413 | 0.9940 | **0.9669** | 0.9209 | 0.9515 | **0.9359** |
| StackVAE+Fixed Graph | 0.9616 | 0.9472 | 0.9543 | 0.8932 | 0.9315 | 0.9119 | 0.7704 | 0.9825 | 0.8636 |
| StackVAE | 0.9605 | 0.9469 | **0.9536** | 0.8948 | 0.9315 | **0.9128** | 0.8999 | 0.9955 | **0.9453** |
| StackVAE-G | 0.9561 | 0.9538 | **0.9550** | 0.9014 | 0.9310 | **0.9160** | 0.9172 | 0.9955 | **0.9547** |

**Table 2**
Hyper-parameters settings.

| Datasets | StackVAE | | StackVAE-G | | | |
|---|---|---|---|---|---|---|
| | $l$ | $m$ | $\gamma$ | $k$ | $\alpha$ | $\lambda$ |
| SMD | 40 | 20 | 0.5 | 10 | 2.0 | 1.0 |
| SMAP, MSL | 100 | 20 | 0.5 | 15 | 2.0 | 1.0 |

- *Fusion ratio*: The parameter $\gamma$ in (Eq. (3)) controls the ratio of fused mid-latent feature $\widetilde{A} \times H_1$ in $H_2$. We search it in the range $[0.1, 1.0]$ with 0.1 as the step size. The detection performance fluctuates mildly and gets the best performance around 0.5, which indicates that the fusion process among channels is beneficial while the original features of each channel also need to be protected.
- *Amplifier*: we use $\alpha$ to represent the amplifier parameter as in Eq. (6). The larger $\alpha$ is, the more relevant neighbours can be found. If the amplifier is set to be 0, apparently the learned graph structure $\widetilde{A}$ will be an identity matrix $\mathbf{I}$ with no neighbours found. And this hyper-parameter is insensitive because as the graph learning loss converges, only the highly-correlated channels are still connected. The larger amplifier provides more neighbour candidates and the graph learning loss decides which can stay. We search it in the range $[1.0, 2.5]$ with 0.1 as the step size, and get the best structure learning performance at 2.0.
- *Top-k*: to ensure the sparsity of the graph, $k$ is set to mask the other elements in a row of $\widetilde{A}$ except the top-k elements. When setting the value of $k$, we need to make sure that $k$ is comparable with the size of the largest cluster of nodes, so that the interrelation structure learned can reveal the complete information. $k$ is selected in the set $\{5, 10, 15, 20\}$ and differs in different datasets.
- *Hyper-parameter for loss function*: we set a hyper-parameter $\lambda$ to balance the two parts of StackVAE-G's loss function. We search it in the range $[0.5, 2.0]$ with 0.5 as the step size.

Table 2 shows the hyper-parameters selected for different datasets. Note that besides the above four hyper-parameters, StackVAE-G has another two — the window size $l$ and the dimension of latent space $m$. StackVAE and StackVAE-G share the parameters $l$ and $m$. Note that besides the window size $l$, $k$ varies in different datasets. A larger $k$ in SMAP and MSL means the size of the largest node cluster is greater than that in SMD, thus more interrelated neighbour candidates are provided for structure learning.

The other parameters for model training are listed in Table 3. They are shared by both StackVAE and StackVAE-G.

*4.4. Analysis on anomaly detection performance*

Table 1 shows the detailed performance for all the selected approaches on the three public datasets, including the strong baselines

**Table 3**
The other parameters.

| Other parameters for training | |
|---|---|
| Optimizer | Adam (Kingma and Ba, 2015) |
| Learning rate | 1e−3 |
| Learning rate decay | 0.8 |
| Weight decay | 1e−3 |
| Gradient clip value | 12.0 |
| Epochs | 256 |
| Batch size | 64 |

**Table 4**
Average performance of StackVAE-G (±standard deviation) on each dataset.

| Datasets | Precision | Recall | F1-score |
|---|---|---|---|
| SMD | 0.9541(0.0015) | 0.9544(0.0010) | 0.9542(0.0007) |
| SMAP | 0.8983(0.0022) | 0.9310(0.0000) | 0.9143(0.0011) |
| MSL | 0.9084(0.0053) | 0.9955(0.0000) | 0.9500(0.0029) |

(e.g., OmniAnomaly, USAD, ITAD, THOC and Anomaly Transformer). We can see that our proposed StackVAE achieves comparable performance to the newly-proposed state-of-the-art Anomaly Transformer and outperforms the others with considerable improvements.

OC-SVM, IF, DeepSVDD and DAGMM present the relatively lower performance because the temporal information within series data is ignored.

As for the StackVAE model variations, StackVAE-G with the graph neural network module shows a comparable or slightly better detection result compared with StackVAE, the one without the graph learning module. Moreover, StackVAE-G also has a better detection result compared with the StackVAE+Fixed Graph method, which reveals that StackVAE-G successfully models the graph structure in the VAE generative modelling.

In addition, to demonstrate the stable detection performance of StackVAE-G, we repeat the training and testing procedures for 5 times individually and report the average performance metrics with the corresponding standard deviations. From Table 4, we can see the average performance is quite close to the best shown in Table 1 and the standard deviations are quite small.

*4.5. Analysis on computation and memory efficiency*

We first report the size of StackVAE-G and other three strong, AE-based baselines, OmniAnomaly, USAD and Anomaly Transformer with corresponding window size $l$ in Table 5. The first row for each model in Table 5 represents the model size with the optimal $l$. Notably, our model size is much smaller than OmniAnomaly and Anomaly Transformer (step-wise) with the same window size. Moreover, StackVAE-G

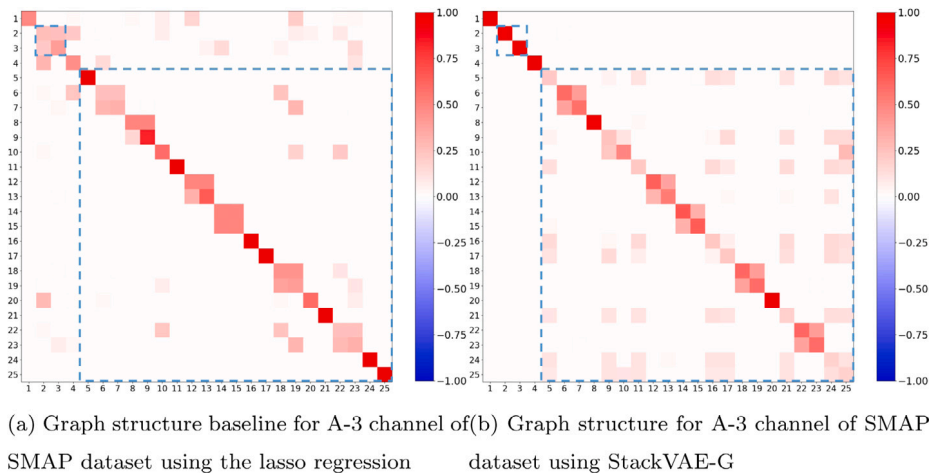

(a) Graph structure baseline for A-3 channel of SMAP dataset using the lasso regression  (b) Graph structure for A-3 channel of SMAP dataset using StackVAE-G with larger window size $l$ holds a comparable amount of parameters with the optimal USAD under $l = 5$.

**Fig. 4.** Comparison between the graph structures learned by the lasso regression and StackVAE-G. The Row (Column) *n* is corresponding to channel *n* of the multivariate time-series data. The coloured pixels in Row (Column) *n* represents the neighbours of channel *n*. The lasso regression fails to learn accurate graph structure due to wrong connection between channel 2,3 and failures among zero-composed channels 5, 11, 16, 17, 21, 24 and 25. StackVAE-G avoids these failures and mistakes, revealing intuitive clustering information among channels.

**Table 5**
Amount of parameters (window size $l$).

|                     | SMD         | SMAP        | MSL         |
|---------------------|-------------|-------------|-------------|
| OmniAnomaly         | 2.61M(100)  | 2.57M(100)  | 2.65M(100)  |
| USAD                | 0.07M(5)    | 0.03M(5)    | 0.15M(5)    |
|                     | 27.23M(100) | 12.67M(100) | 58.81M(100) |
| Anomaly-Transformer | 4.83M(100)  | 4.80M(100)  | 4.86M(100)  |
| **StackVAE-G**      | 0.11M(40)   | 0.15M(100)  | 0.15M(100)  |
|                     | 0.15M(100)  | –           | –           |

**Table 6**
Training time per epoch (min).

| Methods             | SMD    | SMAP   | MSL    |
|---------------------|--------|--------|--------|
| OmniAnomaly         | 87     | 48     | 11     |
| USAD                | 0.6    | 0.08   | 0.03   |
| Anomaly-Transformer | 0.8    | 0.10   | 0.09   |
| **StackVAE-G**      | **0.127** | **0.013** | **0.012** |

Also, we raise USAD as an example to explain why the block-wise models are hard to scale to capture the long-term dependencies, which is easy to achieve for StackVAE-G. To clearly see the model size increment, we additionally report the size of USAD and StackVAE-G under $l = 100$. For USAD, the input and output dimension of each linear model layer are proportional to $N \times l$. Thus the parameter amount of a linear layer is proportional to $N^2 \times l^2$, which explains the model size explosion when $l$ grows larger. On the contrary, under the channel-level weight-sharing scheme, the model size of StackVAE is irrelevant to $N$ (since $N$ collapses to 1), which enables us to catch long-term temporal dependencies with larger $l$.

We also report the average time cost per training epoch on each dataset for the three models in Table 6. (The training time of OmniAnomaly is cited from (Audibert et al., 2020).) As can be seen, StackVAE-G is much more time-efficient compared with the other three baseline models. All methods are trained on one NVIDIA GeForce GTX 1080 Ti. Our stacking block-wise model framework brings small model size and promising training efficiency.

*4.6. Analysis on graph structure and interpretability*

In Fig. 4, we plot the heat maps of the adjacency matrices learned from the A-3 instance of the SMAP dataset to show the graph structure.

The left subplot represents the graph structure learned by the lasso regression (i.e., no graph learning module), while the result of StackVAE-G is shown on the right. We focus on finding the highly-correlated neighbours for each channel and sharing their normal patterns within each group. We can see that both methods discover the similarities between channel (6, 7), (18, 19) and (22, 23), while StackVAE-G succeeds in finding the most related neighbours (e.g., the all-zeros channels 5, 11, 16, 17, 21, 24, and 25, as illustrated in Fig. 1), so as to give a reliable and accurate graph structure. Moreover, in the graph structure of the lasso regression, some irrelevant channels are connected. For example, channels 2 and 3 have totally different curves. We also give an ablation study of the StackVAE-G by removing the graph learning loss in Eq. (8). The results are shown in Fig. 5. We can see that without supervision on graph learning, the graph structure learned is completely uninterpretable and cannot match the curves shown in Fig. 1. It implies that the graph learning loss indeed helps StackVAE-G to correctly learn the stable and reliable graph structure.

Importantly, we recommend the user or the operator take the following steps to interpret the detected results using our model. First, run the model to test the coming time series streams and decide whether an anomaly occurs. When an anomaly is reported, the operator can check the learned graph output adjacency matrix to get the interrelated neighbours of each channel. No matter with or without the prior knowledge for the certain problem, the human operator can easily perform the failure diagnosis to see whether it is a sensor clerical error or a systematic error (illustrated in Fig. 6), which corresponds to a single-channel anomaly and a structural multi-channel anomaly respectively. More specifically, a systematic error occurs when all neighbour channels exhibit anomalous behaviour, whilst a sensor clerical error occurs when there is only one channel showing anomalous behaviour with its neighbour channels remain their normal patterns. This kind of information is highly valuable for system failure diagnosis.

## 5. Conclusion and discussion

In this paper, we propose a stacking VAE model with graph neural networks for efficient and interpretable multivariate time-series anomaly detection. Specifically, we propose a block-wise reconstruction framework with a weight-sharing technique to exploit the channel-level similarities within multivariate time series data, which can significantly reduce the computation and memory cost while maintaining high detection performance.

To exploit the interrelation structural information among the multiple channels, our model learns a sparse adjacency matrix using a graph learning module for all the channels in time-series data. Using the graph learning loss, the graph learning module learns the stable graph structure providing valuable information for reliable latent code inference and fault diagnosis applications.

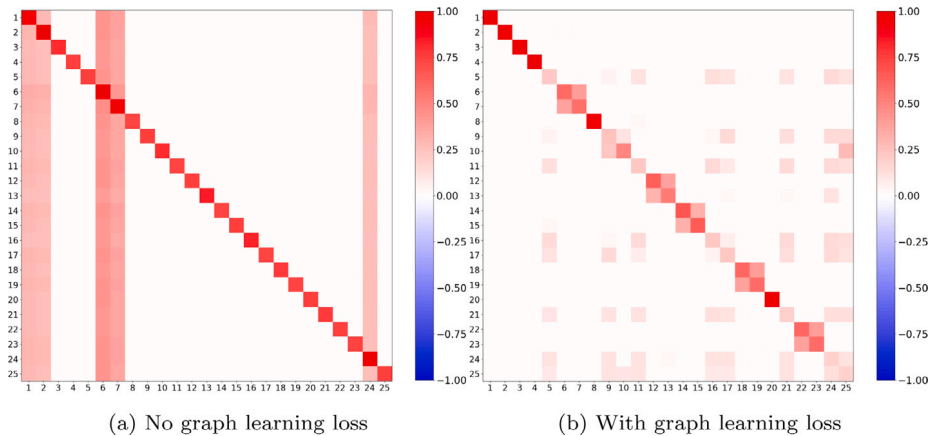

(a) No graph learning loss     (b) With graph learning loss

**Fig. 5.** Comparison between the graph structures learned without and with graph learning loss. Without graph learning loss, StackVAE-G fails to learn meaningful graph structure, indicating transferring the graph learning module from prediction tasks (Wu et al., 2020) directly does not work under stacking block-wise reconstruction framework. graph learning loss ensures that StackVAE-G can learn meaningful and stable graph structure.

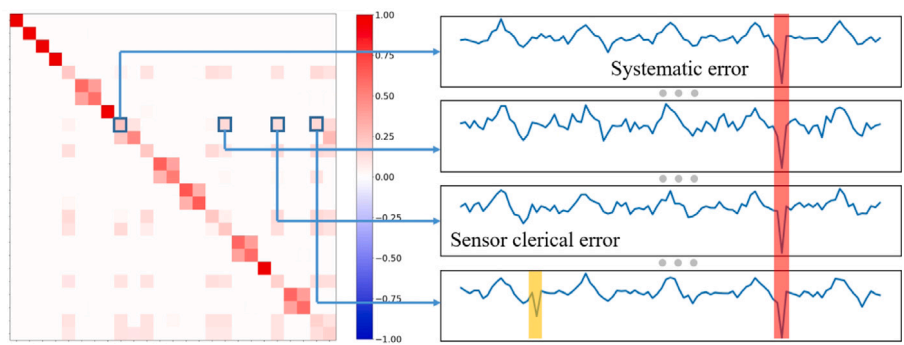

**Fig. 6.** The adjacency matrix provides neighbour channels as references for interpreting the system diagnosis results.

We conduct experiments on three commonly-used multivariate time-series anomaly detection datasets, SMD, SMAP, and MSL. Our proposed model achieves comparable (even better) the latest strong baselines on detection performance. Furthermore, we also show that our model is memory and computing resource-efficient making it suitable for deployment in resource-limited environments like edge devices that are widely used in IoT scenarios.

Lastly, we would like to point out that the interrelation structure we study in this paper is linear (See $\mathcal{L}_{\text{Graph}}$ in Eq. (8)), since the designed graph learning objective $\mathcal{L}_{\text{Graph}}$ is unable to describe the relationship between two variables like $y = x^2$ and other non-linear relationships. Nevertheless, we find the linear interrelation structure captured in our model can already significantly benefit the multivariate time series anomaly detection tasks on both the performance and interpretability perspectives. We leave the capturing of nonlinear interrelation structure as future work.

**Declaration of competing interest**

The authors declare that they have no known competing financial interests or personal relationships that could have appeared to influence the work reported in this paper.

**Acknowledgements**

This work was supported by the NSFC Projects (Nos. U19A2081, U19B2034, U1811461, 62061136001, 61621136008, 62076147), National Key Research and Development Program of China (Nos. 2020AAA0106000, 2020AAA0104304, 2020AAA0106302), the major key project of PCL (No. PCL2021A12), Tsinghua-Huawei Joint Research Program, a grant from Tsinghua Institute for Guo Qiang, Beijing Academy of Artificial Intelligence (BAAI), and the NVIDIA NVAIL Program with GPU/DGX Acceleration.